\documentclass[10pt,twocolumn,letterpaper]{article}

\usepackage[pagenumbers]{cvpr} 

\usepackage{xspace}
\def\ie{\emph{i.e}.\xspace}
\def\eg{\emph{e.g}.\xspace}
\usepackage{booktabs}
\usepackage[dvipsnames]{xcolor}

\definecolor{cvprblue}{rgb}{0.21,0.49,0.74}
\usepackage[pagebackref,breaklinks,colorlinks,citecolor=cvprblue]{hyperref}

\def\paperID{*****} 
\def\confName{CVPR}
\def\confYear{2024}

\title{Solution for SMART-101 Challenge of CVPR Multi-modal Algorithmic Reasoning Task 2024}

\author{Jinwoo Ahn\textsuperscript{\rm 1}$^*$, Junhyeok Park\textsuperscript{\rm 1}$^*$, Min-Jun Kim\textsuperscript{\rm 3}, Kang-Hyeon Kim\textsuperscript{\rm 2}, So-Yeong Sohn\textsuperscript{\rm 1},\\
Yun-Ji Lee\textsuperscript{\rm 3}, Du-Seong Chang\textsuperscript{\rm 3}, Yu-Jung Heo\textsuperscript{\rm 3}$^\dagger$,  Eun-Sol Kim\textsuperscript{\rm 1,2}$^\dagger$\\
\textsuperscript{\rm 1}Department of Artificial Intelligence Application, Hanyang University\\
\textsuperscript{\rm 2}Department of Artificial Intelligence, Hanyang University \\
\textsuperscript{\rm 3}KT Corporation \\
{\tt\small \{jinwooahn, bear1123, khflower, soyeong2, eunsolkim\}@hanyang.ac.kr} \\
{\tt\small \{min\_jun.kim, lee.yunji, dschang, yj.heo\}@kt.com }
}

\begin{document}
\maketitle
\def\thefootnote{*}\footnotetext{These authors contributed equally to this work.}
\def\thefootnote{$\dagger$}\footnotetext{Corresponding author}
\begin{abstract}
In this paper, the solution of $\text{HYU\_MLLAB\_KT Team}$ to the Multimodal Algorithmic Reasoning Task: SMART-101 CVPR 2024 Challenge is presented. Beyond conventional visual question-answering problems, the SMART-101 challenge aims to achieve human-level multimodal understanding by tackling complex visio-linguistic puzzles designed for children in the 6-8 age group. To solve this problem, we suggest two main ideas. First, to utilize the reasoning ability of a large-scale language model (LLM), the given visual cues (images) are grounded in the text modality. For this purpose, we generate highly detailed text captions that describe the context of the image and use these captions as input for the LLM. Second, due to the nature of puzzle images, which often contain various geometric visual patterns, we utilize an object detection algorithm to ensure these patterns are not overlooked in the captioning process. We employed the SAM algorithm, which can detect various-size objects, to capture the visual features of these geometric patterns and used this information as input for the LLM.  Under the puzzle split configuration, we achieved an option selection accuracy ${O_{acc}}$ of 29.5 on the test set and a weighted option selection accuracy (WOSA) of 27.1 on the challenge set.
\end{abstract}   
\section{Introduction}
\label{sec:1_Introduction}

Visual Question Answering (VQA) task requires the model to answer natural language questions about input images. With the growth of multimodal large language models (MLLMs), these models have demonstrated significant performance in understanding the complex context of multimodal inputs (\ie, a given image and question) and generating an appropriate response to the question. 
Recently, several benchmarks, including ScienceQA~\cite{ScienceQA} and MMMU~\cite{MMMU}, have been introduced to assess the multimodal reasoning capabilities in specialized fields. These benchmarks aim to evaluate the expert knowledge of large language models (LLMs) through subject-specific questions spanning diverse areas such as mathematics, science, medicine, business, and the arts.
On the other hand, the SMART-101 challenge~\cite{SMART101_Benchmark} evaluates the abstraction, deduction, and generalization abilities of neural networks in solving visio-linguistic puzzles designed specifically for children in the 6–8 age group. The SMART-101 dataset consists of 101 unique root puzzles that require a mix of various skills such as counting, spatial localization, and mathematical ability. Therefore, we required fine-grained perception ability, as the MLLM must handle complex synthetic images in diagram form rather than real-world visual scenes and complex reasoning ability is necessary due to the need for difficult skills and their combinations not typically required in traditional VQA tasks.


To tackle the SMART-101 challenge, we propose a new instruction-tuned vision-language model with two novel ideas. First, to utilize the reasoning ability of pre-trained MLLMs, the given visual cues (images) are grounded in the text modality. We generate highly detailed text captions that describe the context of the image and use these captions as input for the pre-trained MLLMs. Second, due to the nature of puzzle images, which often contain complex diagrammatic visual patterns, we utilize an object detection algorithm to ensure these patterns are not overlooked in the captioning process. To tackle this problem, the Segmentation Anything Model (SAM) algorithm is introduced to capture the complex visual feature, and the visual features from the SAM are used as another input of the pre-trained MLLMs. Through these methods, we achieved a 27.11 WOSA score on the challenge split and qualitatively validated the effectiveness of our proposed approach.

\section{Related work}
\label{sec:2_Related_work}
Before proposing our method, we briefly review a few prior methods for understanding the multimodal large language models and the importance of visual understanding in solving puzzle VQA datasets.
\subsection{Large language model}

In recent years, large language models (LLMs) have been rapidly evolving in the field of natural language processing.  LLMs are capable of learning new tasks efficiently with only a few examples, showing few-shot and zero-shot learning aptitude~\cite{brown2020language}.
However, despite their flexibility, LLMs encounter significant limitations, especially in their reasoning and computational abilities. For instance, advanced open API models (\eg GPT-4) have difficulty solving even addition problems with large numbers in a zero-shot setting without external tools~\cite{mcleish2024transformers}. This shows that while LLMs excel at learning language patterns, they have limitations in fully understanding and applying the algorithmic nature of mathematical operations. Furthermore, the performance of LLMs is below expectations in tasks that require complex logical reasoning~\cite{imani2023mathprompter}.

\subsection{Multimodal large language model}

Recently, multimodal large language models (MLLMs) such as Gemini~\cite{geminiteam2024gemini}, BLIP-2~\cite{li2023blip2}, and LLaVA~\cite{liu2023llava} have gained attention for incorporating different modalities, going beyond large language models (LLMs). These MLLMs process information from multiple modalities, making their thinking more human-like than LLMs which specialize in text-based tasks.
However, MLLMs still have limitations, such as a lack of accuracy in alignment between modalities, text-biased inference, and imbalanced cross-modal interactions that rely only on some attention heads~\cite{cao2020behind}. To solve these problems, researchers are trying a variety of approaches. BLIP-2 proposed a new pre-training method that explicitly models the alignment of image-text pairs. Flamingo~\cite{alayrac2022flamingo} uses a gated cross-attention mechanism to effectively model the interaction between visual and text representation. Our backbone model, InstructBLIP~\cite{InstructBLIP}, enhanced instruction following ability through supervised fine-tuning. While these studies have been improving MLLMs' performance and image inference, MLLMs still need some improvement when it comes to solving visual math problems~\cite{zhang2024mathverse}.

\subsection{Text-enhancement in MLLMs}

Various approaches are adopted to extract visual information from images and convert it into textual representations to aid the model in comprehending visual features. Zheng et al.~\cite{zheng2024cogview3} enhance the input text by feeding image-caption pairs into VLM to generate visual questions and answers that describe the image. Then, they are used to create the final caption, which enhances the quality of diffusion-based text-to-image generation. J. Park et al.~\cite{Park2023LocalizedSK} utilize BLIP-2 and other descriptors to extract global and local information from images to enable VLM to extract local information more effectively. They then instruct LLM to generate a question-answer-rationale (QAR) triplet as a knowledge representation, which describes an image in text using this information. In the end, they refine the BLIP-2 model by selecting only the correct information via the critic model.

\subsection{Visual understanding}


A puzzle image is a compressed representation of information. They are designed to make it easy for humans to understand the information, which makes them difficult for computers to understand~\cite{kafle2018dvqa}. While a typical VQA task focuses on figuring out the overall meaning of an image, understanding a puzzle image requires more precise analysis and interpretation of visual elements. The model needs to be able to distinguish between the foreground and background of the image, and even small changes in the content, such as what is the larger object. Therefore, unlike traditional VQA tasks, we require another perspective of visual understanding skills to solve the puzzle image.


\section{Method}
\label{sec:3_Method}


\begin{figure*}[t]
    \centering
    \includegraphics[width=\linewidth]{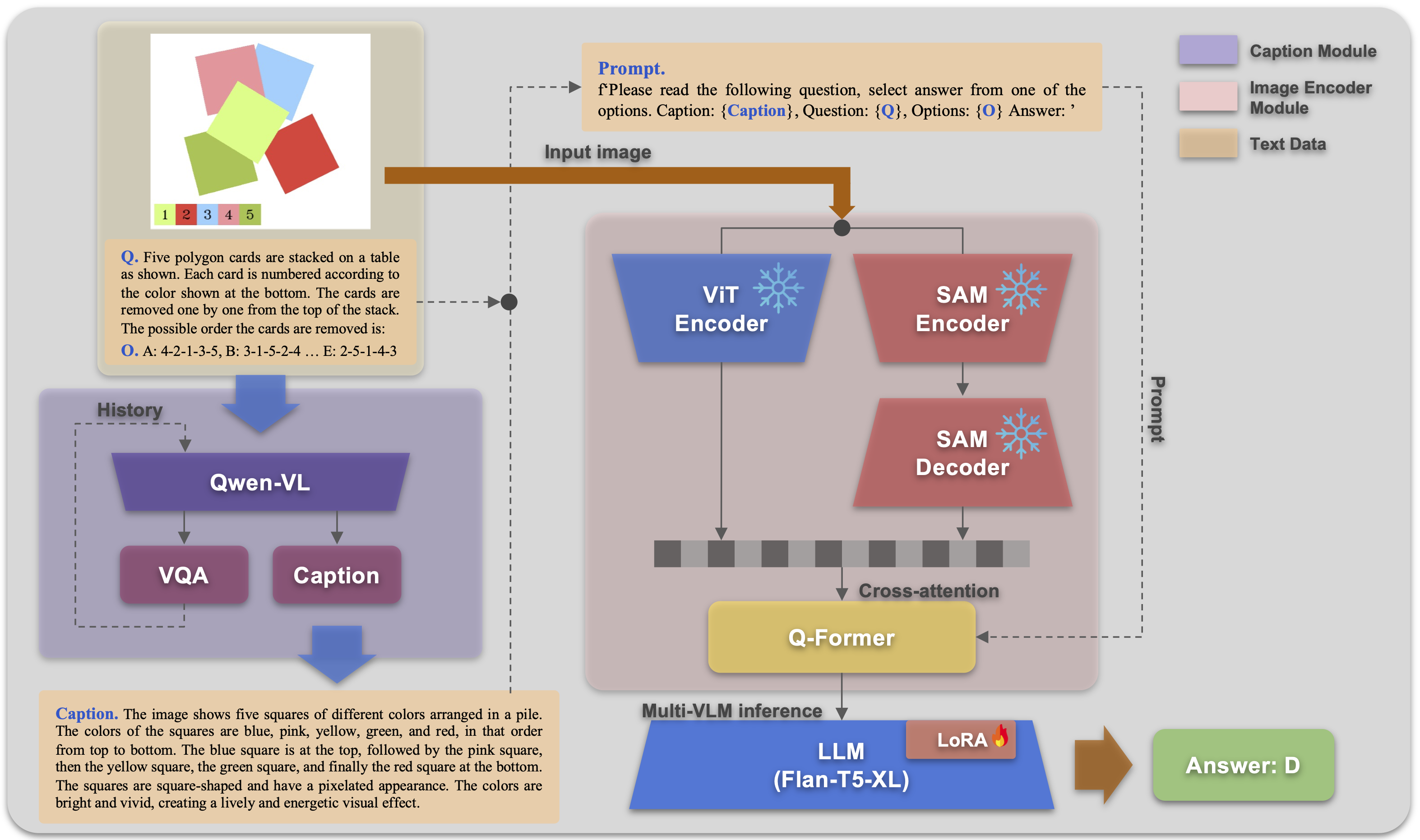}
    \caption{\textbf{Overall framework of proposed pipeline.} In the framework, we extract image captions using a two-stage mechanism. To enhance the quality of the caption, we first generate three sets of visual question and answer (VQA). The results of these VQA generations are then used as history and included as additional prompts when generating the caption of the image. The generated caption, along with the question, is used as a prompt for the backbone model, InstructBLIP. We enhanced the visual understanding ability by concatenating features from ViT and SAM. Finally, the image and text embeddings are processed through Q-Former and LLM with a specific ensemble strategy by classifying puzzle categories.}
    \label{figure:fig_main_arch}
\end{figure*}
The overall architecture of our approach is illustrated in Figure \ref{figure:fig_main_arch}, consisting of three components: (i) Text enhancement module, (ii) Vision enhancement module, and (iii) Multi-VLM training and inference module. The selection of the backbone VLM is not restricted; here, we selected the InstructBLP~\cite{InstructBLIP}, a well-known general-purpose VLM, as our baseline.

\subsection{Text enhancement module}

\begin{figure*}[t]
    \centering
    \includegraphics[width=0.85\linewidth]{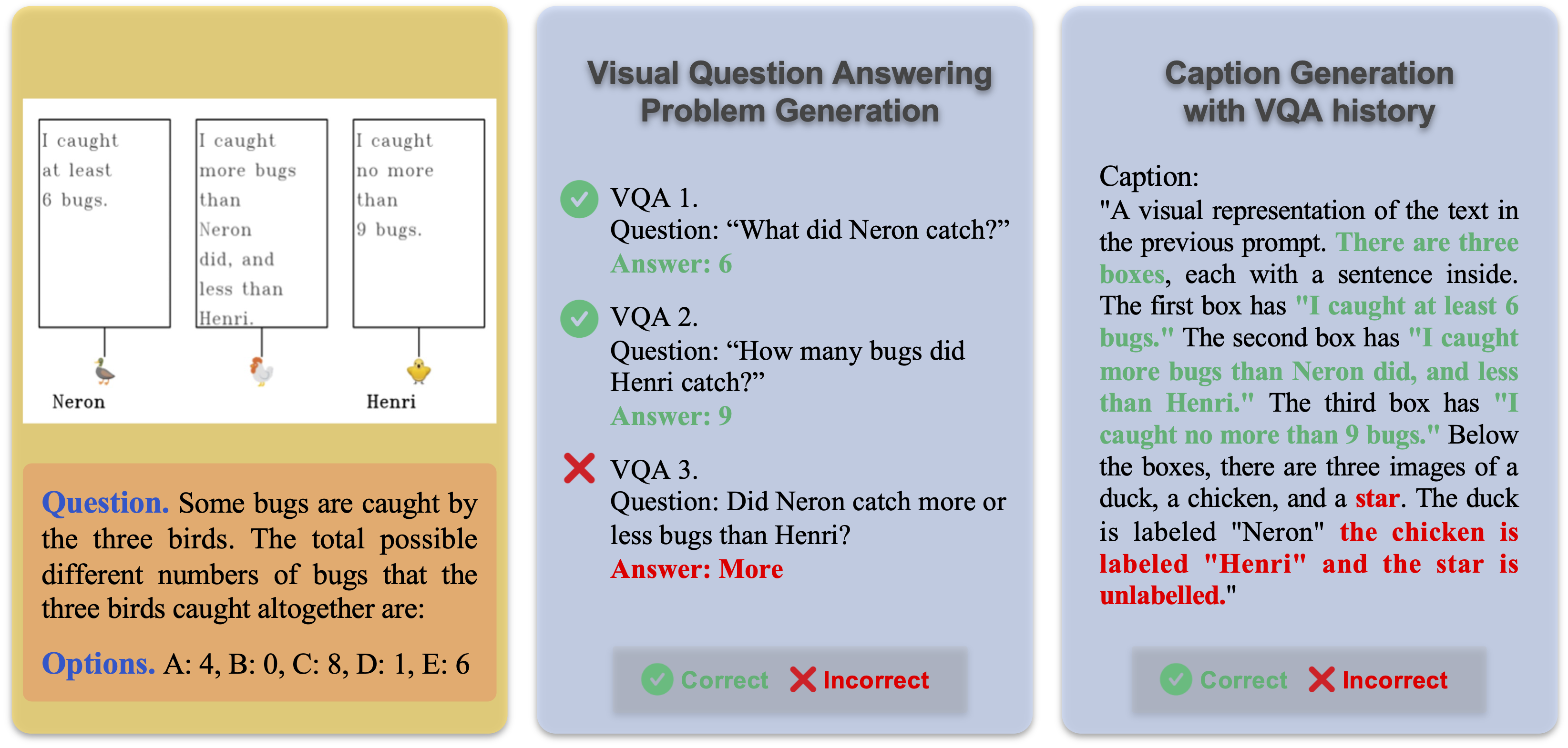}
    \caption{\textbf{Result of text enhancement module.} The left box is the target puzzle to augment text information through Qwen-VL-Chat, and the middle box is the generated visual question-answer pairs. The right box is the result of generating captions using VQA pairs as history.}
    \label{figure:fig_caption_module}
\end{figure*}

Recent studies have shown that leveraging the text-based reasoning ability of large-scale language models (LLMs) is effective even for image-based question answering tasks, where questions and answers are derived from given images. These studies adopt a strategy of transforming the content of images into a form that LLMs can understand (language grounding) rather than directly utilizing visual features, and then feeding this transformed content into the LLMs.

In this study, we also chose to utilize LLMs by converting the content of images into text captions. However, the images used in the SMART-101 challenge are characterized by many geometric shapes and are specialized for puzzles, making them unsuitable for general image captioning algorithms. To tackle this problem, we used the Qwen-VL \cite{Qwen-VL} as the image captioning method, because the Qwen-VL is trained on DocVQA \cite{DocVQA} and ChartQA \cite{ChartQA} datasets which contain various types of documents and charts. 

Furthermore, we adopted a two-step strategy where we first generate question-and-answer pairs about the image, and then create captions based on these pairs. This approach ensures that the captions encompass various levels of semantics present in the image. The text generation results of Qwen-VL are illustrated in Figure \ref{figure:fig_caption_module}.


\subsection{Vision enhancement module}
Although we can achieve high performance in visual question answering (VQA) by converting the meaning of images into text as much as possible and leveraging the reasoning ability of LLMs, we also utilized object detection-based image features to prevent potential information loss from language grounding. However, since the images used in the SMART-101 challenge differ in nature from natural images, we used the Segmentation Anything (SAM) algorithm \cite{SAM}, which can extract geometric patterns from the images, rather than relying on general object detection algorithms.

Since our vision enhancement module was trained specifically for segmentation, enabling the extraction of fine-grained, object-level features from the image. These features are integrated with the features of Vision Transformer (ViT) from instructBLIP and directed into the q-former's cross-attention mechanism to enrich the available features for visual reasoning.

\subsection{Training with additional datasets}
Although InstructBLIP shows robust instruction following capability learned by multi-task finetuning, the model is mostly trained on natural image-based datasets for general-purpose VLM. Furthermore, several crucial visual reasoning datasets were selected as held-out datasets and not used in the training phase for InstructBLIP. Therefore, we supplement additional datasets to foster visual reasoning ability (\eg, mathematical reasoning, geometric visual reasoning) for the baseline model on complex synthetic puzzle images used in the smart-101 challenge. The details of the additional datasets that we choose are described in Sec \ref{details_addidata}.

\subsection{Multi-VLM training and inference}
\begin{figure}[t]
    \centering
    \includegraphics[width=\columnwidth]
    {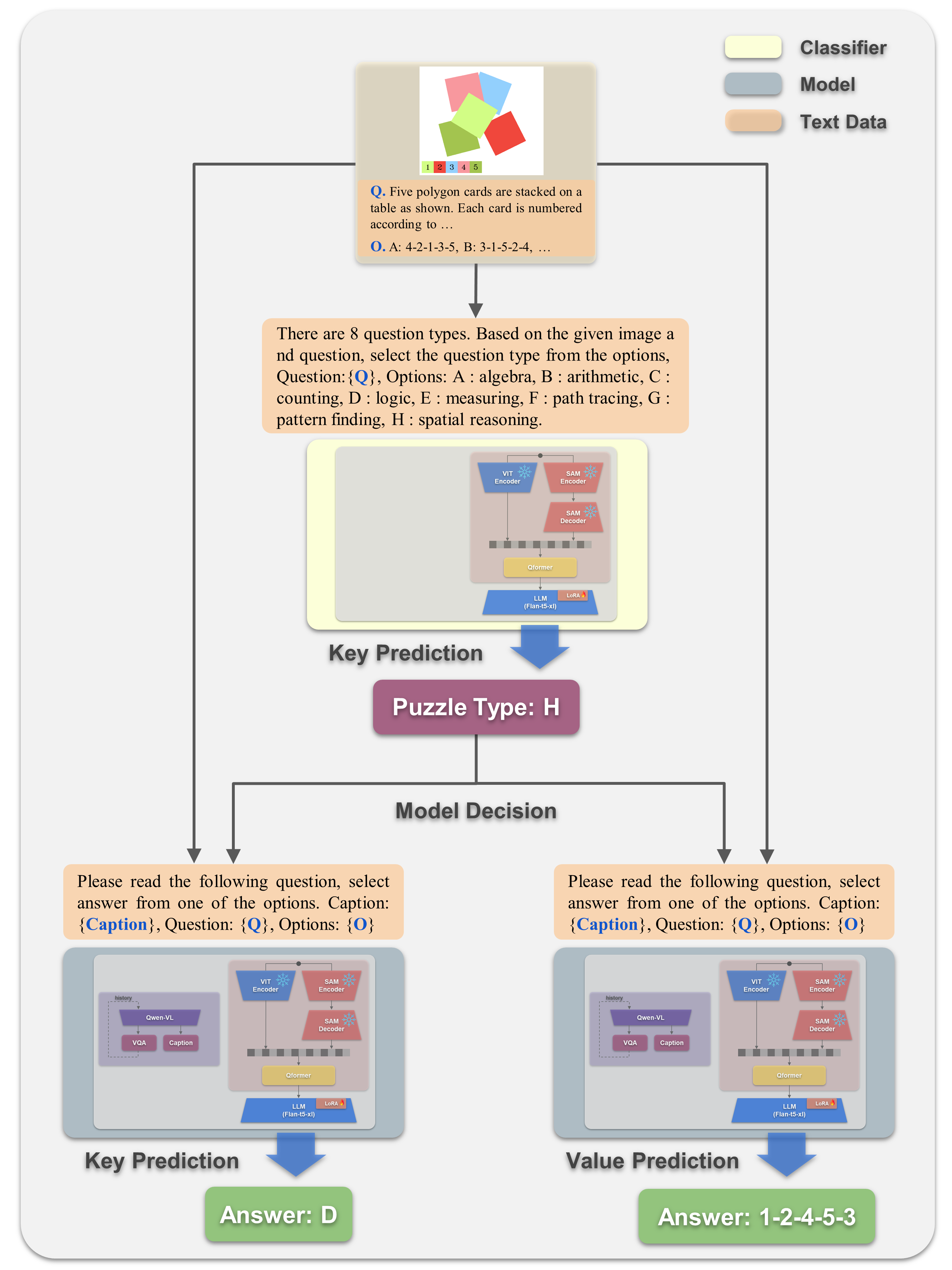}
    \caption{\textbf{Inference workflow for Multi-VLM.} A zero-shot classifier determines the puzzle category. Based on the classified puzzle type, either the key prediction model or the value prediction model is selected, each of which is specifically trained for the corresponding answer type.}
    \label{figure:fig_infer_arch}
\end{figure}

To achieve optimal predictive performance, two distinct models were trained in parallel according to the type of puzzles: (i) For puzzles categorized under logic, counting, spatial reasoning, path tracing, and pattern finding, a model was trained to predict the option ``key". (ii) For puzzles that deal with arithmetic, measurement, and algebra, a model was trained to predict the answer ``value". This approach addresses the performance variability across puzzle types, enabling more accurate predictions by utilizing the most suitable model for each puzzle category. During inference, a zero-shot classifier identifies the puzzle type and selects the appropriate model, either the key model or the value model. The zero-shot classifier, designed to improve generalization to new puzzles, is based on the key model, which exhibited high classification accuracy (0.36 for eight puzzle types and 0.8 for key/value types). The inference workflow and the prompts employed for classification are illustrated in Figure \ref{figure:fig_infer_arch}.
\section{Experiment}
\label{sec:4_Experiment}

\subsection{Dataset}
\subsubsection{Smart-101}
In this challenge, we mainly use the SMART-101 dataset \cite{SMART101_Benchmark}. It is built from 101 distinct children's puzzles, referred to as root puzzles, on paper. And based on these, there are 2,000 instance puzzles, each of which has a similar format to root puzzles. Therefore, the dataset consists of a total of 202,000 puzzles.

\subsubsection{Additional datasets}\label{details_addidata}
To enhance the multimodal reasoning capabilities of the VLM for various types of diagram images, we utilized the following additional datasets in the training process.
\begin{itemize}
    \item MathVista~\cite{mathvista}, Math-Vision~\cite{mathvision}, and MathVerse~\cite{MathVerse} for multimodal mathematical reasoning
    \item RAVEN~\cite{raven} and IconQA~\cite{iconqa} for relational and geometric visual reasoning on abstract diagram images
    \item ScienceQA~\cite{ScienceQA} and MMMU~\cite{MMMU} for multi-hop reasoning with subject-specific knowledge such as natural science and social science
    \item MMBench~\cite{mmbench} and  MMStar~\cite{mmstar} for fine-grained perception and reasoning ability
\end{itemize}

Following the QA type of the smart-101 dataset, we sampled only the multiple-choice QA types from the aforementioned datasets. The total number of additional training data instances is 118,011.

\subsection{Evaluation metrics}
According to the evaluation guideline of challenge, we employ the zero-shot generalization setting outlined in the original paper, referred to as Puzzle Split. This entails assessing novel, previously unobserved root puzzle instances that demand the same foundational skills as those in the training set. Basically, when we evaluate the model at the test set in the benchmark, we only use the option selection accuracy ($\mathcal{O}_{acc}$) that measures the frequency with which the correction option is selected from one of the five answer options by a model. But in the challenge test phase, considering the diverse difficulty of the puzzles, we weigh the puzzle when scoring the performance. Thus, to compute the final score for a model, we use a weighted option selection accuracy (WOSA) defined in formula \ref{formula:for_wosa} where ${w_{i}}$ represents the weight of each puzzle in the test set, where $\mathcal{O}_{acc}^{i}$ is 1 if the answer is correct, and 0 otherwise.
\begin{equation}
  WOSA = 100 \times \frac{{\sum_{i=1}^{N}}w_{i}*\mathcal{O}_{acc}^{i}}{{\sum_{i=1}^{N}}w_{i}}
  \label{formula:for_wosa}
\end{equation}

\subsection{Implementation details}
In our study, the InstructBLIP-Flan-T5-XL \cite{InstructBLIP} model is selected as the pretrained MLLM method. The training was conducted using 4 A100 GPUs, and optimal performance was achieved at approximately 2 epochs. The learning rate was set to 1e-5 with a batch size of 16 per GPU. Furthermore, we performed LoRA fine-tuning for MLLM with a learning rate of 1e-6, focusing on the attention module.

\subsection{Results}



\begin{table}[h]
\centering
\scalebox{0.90}{
{\small
\begin{tabular}{lccc}

\toprule
Method & Text-WOSA & VL-WOSA & Total-WOSA\\
\midrule
IB + MLP classifier & 16.49 & 25.90 & 22.03 \\
IB + Extra datasets & 23.71 & 25.18 & 24.58 \\
IB + SAM & 35.05 & 20.86 & 26.69 \\
\midrule
IB + All (Ours) & 30.93 & 24.46 & \textbf{27.11} \\
\bottomrule
\end{tabular}
}
}
\caption{\textbf{Test accuracy achieved by the suggested method during the competition} \textbf{IB} represents InstructBLIP-Flan-T5-XL baseline model. \textbf{IB + All} represents the suggested method including Qwen-VL caption, SAM, Multi-VLM inference, and training with additional datasets.} 
\label{tab:challenge_result}
\end{table}





Table \ref{tab:challenge_result} represents the test accuracy achieved by our model during the competition. As explained above, we select the InstructBLIP-Flan-T5-XL algorithm as the base MLLM architecture and show the performances when combining this model with the ideas described in Section \ref{sec:3_Method}. We show that combining the two main ideas (using highly-detailed captions and object-oriented visual features) with the two training details (training with additional datasets and multi-VLM inference) results in high performance.



\section{Conclusion}
\label{sec:5_Conclusion}

In this paper, we present a novel method to solve complex puzzle problems by enhancing the reasoning performance of the pre-trained multimodal large-scale language models (MLLMs). Specifically, we suggest leveraging highly detailed captions about the given images and object-oriented visual representations followed by additional training with other datasets. During the competition, we showed that combining the ideas results in high performances.

{
    \small
    \bibliographystyle{ieeenat_fullname}
    \bibliography{main}
}


\end{document}